\pdfoutput=1

\documentclass[11pt]{article}

\usepackage{acl}
\usepackage{float}
\usepackage{times}
\usepackage{latexsym}

\usepackage[T1]{fontenc}

\usepackage[utf8]{inputenc}

\usepackage{microtype}

\usepackage{csquotes}
\usepackage{url}
\usepackage{graphicx}
\usepackage{multicol,array,booktabs,multirow}
\usepackage{xcolor}
\usepackage{subfigure}
\usepackage{lipsum}
\usepackage{enumerate}
\usepackage{tikz-dependency}
\usepackage{pbox}

\newcommand{\parados}[1]{\vspace{0.2em}\noindent\textbf{#1}\hspace{0.2em}}
\definecolor{depred}{HTML}{b12019}
\definecolor{depblue}{HTML}{165f77}
\newcommand{{\mcrae}}{\textsc{McRae}}
\newcommand{\norms}{\textsc{Norms}}
\usepackage{microtype}

\title{Assessing the Limits of the Distributional Hypothesis in Semantic Spaces: \\ Trait-based Relational Knowledge and the Impact of Co-occurrences}

\author{Mark Anderson \\
  PIN Caerdydd\\
  Prifysgol Caerdydd\\
  \normalsize{\texttt{AndersonM8@caerdydd.ac.uk}} 
  \\\And
  Jose Camacho-Collados \\
  Cardiff NLP \\
  Cardiff University \\
\normalsize{\texttt{CamachoColladosJ@cardiff.ac.uk}}
  }
\date{}

\begin{document}
\maketitle
\begin{abstract}
The increase in performance in NLP due to the prevalence of distributional models and deep learning has brought with it a reciprocal decrease in interpretability. This has spurred a focus on \textit{what} neural networks \textit{learn} about natural language with less of a focus on \textit{how}. Some work has focused on the data used to develop data-driven models, but typically this line of work aims to highlight issues with the data, e.g. highlighting and offsetting harmful biases. This work contributes to the relatively untrodden path of what is required in data for models to capture meaningful representations of natural language. This entails evaluating how well English and Spanish semantic spaces capture a particular type of relational knowledge, namely the traits associated with concepts (e.g. \textit{bananas}-\textit{yellow}), and exploring the role of co-occurrences in this context. 
\end{abstract}

\section{Introduction}
Vector space models 
have been the main driving force behind 
progress in NLP. Most work in this area, either in the form of static or contextualised embeddings, has been based on co-occurrence statistics and largely driven by the distributional hypothesis \cite{Harris1954DistributionalS,Firth1957ASO}. This has also resulted in these representations seemingly capturing certain relational knowledge, such as word analogies \cite{Mikolov2013LinguisticRI,gittens2017skip}. In this context, \citet{chiang-etal-2020-understanding} found that the ability of word embeddings to evaluate analogies was not greatly impaired by removing  
co-occurrences related to relational pairs. This suggests there are limits to 
how the distributional hypothesis impacts the encoding of relational knowledge. 
We extend this line of work by focusing on the relational knowledge of concepts and traits. We also creep beyond English by translating concept and traits used in one of our datasets into Spanish. 

\parados{Contributions:} \textbf{(1)} We show that there is no impact on the ability of semantic spaces to predict whether a pair of embeddings corresponds to a trait-concept pair or to predict what traits a given concept has when removing co-occurrences of concepts and traits. 
\textbf{(2)} We developed a freely available dataset that can be used for further trait-based relational knowledge analyses for English and Spanish.\footnote{\url{https://github.com/cardiffnlp/trait-concept-datasets}}

\section{Related work}

\parados{What models learn} Evaluation of \textit{neural} semantic spaces has focused on what knowledge they capture with a slew of work showing that some knowledge of analogies can be seen by applying simple transformations \cite{Mikolov2013LinguisticRI,levy-goldberg-2014-linguistic,Arora2016ALV,Paperno2016WhenTW,gittens2017skip,ethayarajh-etal-2019-towards}. Others have investigated what syntactic information neural semantic spaces seem to capture with most showing that they do capture something deeper than surface patters \cite{linzen2016assessing,gulordava-etal-2018-colorless,giulianelli}. However, they fail to  exhaustively capture syntactic phenomena and specifically have been shown to struggle with polarity \cite{futrell2018rnns,jumelet2018do} and certain \textit{filler-gap} dependencies \cite{wilcox2018what,chowdhury2018rnn}. Pretrained language models (PLMs) have been found to capture varying degrees of syntactic information \cite{peters2018dissecting,tenney2019bert,goldberg2019assessing,clark2019does}, however, they have also been shown to struggle to predict the grammaticality of sentences \cite{marvin2018targeted,warstadt2019neural} and seem to depend on fragile heuristics rather than anything deeper \cite{mccoy2019right}.

\parados{Relational knowledge} More specifically with respect to relational knowledge and semantic spaces, for some time now work has shown that semantic spaces could encode certain relational knowledge, e.g. knowledge of the relative positioning of geographical locations \cite{Louwerse2009LanguageEG}. Similarly, \citet{gupta-etal-2015-distributional} found that embeddings capture something of relational knowledge associated with countries and cities, e.g. how countries related to one another with respect to GDP. \citet{rubinstein-etal-2015-well} found that word embeddings captured some taxonomic relational knowledge but fared less well with respect to trait-based relational knowledge. Often analogy completion tasks are used to investigate what sort of relational knowledge a semantic space has captured with early work showing that simple linear transformations were enough to highlight analogies \cite{Mikolov2013EfficientEO,vylomova-etal-2016-take}. This method has drawn some criticism and has been challenged as a robust means of evaluating what relational knowledge models capture \cite{drozd-etal-2016-word,DBLP:conf/naacl/GladkovaDM16,schluter2018word,bouraoui-etal-2018-relation}. Attempts to evaluate what PLMs capture of relational knowledge have also been made, highlighting that these larger, more data-hungry models capture some but not all relational knowledge \cite{Forbes2019DoNL,bouraoui2020inducing}. 

\parados{Patterns in data} However, all the work cited above focuses work focuses on \textit{what} models learn about relational knowledge and not \textit{how}, or rather what are the salient signals in the data used in these techniques that manifest in relational knowledge. Some work has been done in this direction, with \citet{Pardos2020AUM} showing co-occurrences are not necessary in their distributional model of courses to predict similar or related courses. \citet{chiang-etal-2020-understanding} evaluated this finding in neural semantic spaces, finding that the ability of a semantic space to complete analogies isn't impacted when removing co-occurrences

It is important to understand what aspects of the data result in what models learn because without this semblance of interpretability, problematic biases can creep in, e.g. gender biases in Word2Vec \cite{bolukbasi2016man} or in BERT \cite{Bhardwaj2021InvestigatingGB}. Attempts have been made to mitigate certain biases in contexualised word embeddings \cite{Kaneko2021DebiasingPC}, but in order to do so, the biases have to be known. Also, \citet{shwartz-choi-2020-neural} discuss the issue of reporting bias in the data typically used in NLP, where rarer occurrences are more likely to be explicitly mentioned than common ones which results in models that can generalise about under-reported phenomena but not temper the over-reported information. Therefore it is necessary to understand the nature of the data and how it impacts what models capture and how. 

In this work, we aim to expand on the work of \citet{chiang-etal-2020-understanding} in two main ways. First, we do not use analogies and analogy completion to evaluate the impact co-occurrences of concept-traits has on relational knowledge developed in neural semantic spaces, but instead use a dataset of different trait-based relations (e.g. \texttt{is-colour}, \texttt{has-component}) derived from the {\mcrae} and {\norms} feature datasets. This allows us to more directly evaluate the ability of models to predict relational knowledge by casting the evaluation as a simple classification task (both in a multi class and binary class setting). And second, we extend the analysis by looking at Spanish data as well to evaluate whether the results extend beyond English.

\section{Methodology}
The methodology follows five sequential steps: the development of datasets that include concepts and their traits (Section \ref{secdatasets});  the selection and processing of large general-domain corpora (Section \ref{seccorpora}); the transformation of the selected corpora based on the concept-trait datasets to test our hypothesis (Section \ref{secremoving}); training of  word embeddings on the original and adapted corpora (Section \ref{secembeddings}); and finally the evaluation of the embeddings based on the trait-based datasets (Section \ref{secclassifiers}).

\begin{table*}[tbph!]
    \centering
    \footnotesize
    \tabcolsep=.055cm
    \begin{tabular}{llr>{\raggedleft\arraybackslash}p{2.1em}cr}
    \toprule
    & trait type & \multicolumn{1}{c}{N$_C$} && N$_T$ &\multicolumn{1}{c}{Traits} \\
    \midrule
    \parbox[t]{2.5mm}{\multirow{5}{*}{\rotatebox[origin=c]{90}{\textbf{{\mcrae}-EN}}}} & colour  & 148 && 7 & green (32), brown (32), black (24), white (21), red (16), yellow (13), orange (10) \\
    & components  & 110 && 6 & handle (39), legs (19), wheels (14), leaves (14), seeds (13), doors (11) \\
    & materials  & 144 && 4 & metal (79), wood (43), cotton (11), leather (11)\\
    & size \& shape  & 234 && 4 & small (83), large (70), long (44), round (37)\\
    & tactile  & 117 && 7 & heavy (21), soft (19), furry (18), sharp (17), hard (16), juicy (16), slimy (10)\\
    \midrule
    \parbox[t]{2.5mm}{\multirow{5}{*}{\rotatebox[origin=c]{90}{\textbf{{\norms}}}}} & colour  & 133 & (78) & 5 & green (35), brown (32), white (30), black (22), yellow (14) \\
    & components  & 35 & (26) & 2 & handle (25), sugar (10)\\
    & materials  & 94 & (62) & 5 & metal (46), wood (16), water (11), paper (11), bones (10)\\
    & size \& shape  & 242 & (138) & 4 & small (109), large (73), long (31), round (29)\\
    & tactile  & 106 & (70) & 6 & heavy (28), sharp (26), liquid (14), light (13), juicy (13), soft (12)\\
    \midrule
    \parbox[t]{2.5mm}{\multirow{5}{*}{\rotatebox[origin=c]{90}{\textbf{{\mcrae}-ES}}}} & colour  & 140 && 7 &  verde (31), marrón (31), blanco (21), negro (20), rojo (16), amarillo (12), naranja (9)\\
    & components  & 100 && 6 & mango (33), piernas (18), ruedas (14), hojas (14), semillas (11), puertas (10) \\
    & materials  & 131 && 4 & métal (72), madera (38), algodón (11), cuero (10)\\
    & size \& shape  & 216 && 4 & pequeño (75), grande (66), largo (41), redondo (34)\\
    & tactile  & 101 && 6 & pesado (19), suave (19), peludo (17), duro (16), afilado (16), jugoso (14)\\
    \bottomrule
    \end{tabular}
    \caption{Dataset statistics: N$_C$ is the number of concepts, N$_T$ is the number of unique features, {\norms} N$_C$ includes unique count in parenthesis, and the number in parenthesis for traits is the number of concepts with that trait.}
    \label{tab:dateset}
\end{table*}
\subsection{Datasets}\label{secdatasets}
The datasets were based on the {\mcrae} features dataset \cite{McRae2005SemanticFP}. This is a collection of semantics features associated with a large set of concepts (541) generated from features given by human participants. A secondary trait-based dataset was also collated for English based on the {\norms} dataset \cite{Devereux2014TheCF}. This is developed in the same way as {\mcrae} and is partially an extension of that dataset with 638 concepts. We wanted to avoid value judgements (such as \texttt{is-feminine}) and to collate more trait-based relations, that is pairs of words related by an inherent attribute of a concept. 

\parados{\mcrae-EN} The first step in developing the datasets used in this work was to collate certain features into subsets of similar traits. This was done in a partially manual way by splitting data into 5 subsets. Each feature in {\mcrae} has the number of participants who specified that feature for that concept, so initially a frequency cut of 10 was applied to the features. From this set, we observed a number of similar traits that broadly fit into trait categories. A series of simple heuristics were then applied to extract all potential concept-feature pairs for each subset. For some trait types this was trivial with the {\mcrae} dataset, e.g. colour relations could be found using the feature classification in {\mcrae} of  \texttt{visual-colour}. The full details of the heuristics can be seen in Appendix \ref{sec:heuristics}.

This process resulted in 5 trait-based subsets: \textbf{colours}, \textbf{components}, \textbf{materials}, \textbf{size \& shape}, and \textbf{tactile}. From each subset, we removed duplicates (e.g. ambulance has the features \texttt{is-white},  \texttt{is-red}, and \texttt{is-orange} in the colour subset).\footnote{A multi-label version of these subsets are included at \url{https://github.com/cardiffnlp/trait-concept-datasets} for {\mcrae}-EN and {\norms}-EN.}  And from the remaining concept-feature pairs, we cut on 10+ concepts per trait to ensure a suitable number of instances per target in our evaluation. The resulting statistics associated with this dataset can be seen in the top section of Table \ref{tab:dateset}.

\parados{\mcrae-ES} The set of concepts and trait words that occur across all 5 subsets were manually translated. The translators consisted of one native English speaker with some knowledge of Spanish and one native Spanish speaker who is fluent in English. 

As might be expected, issues occurred when undertaking the translation that required judgements to be made. When there was a one to many translation, we used the translation that was \textit{Iberian} if multiple translations were due to regional variants. Otherwise we chose the most common or most canonical. However, we also chose single word alternatives to avoid multiword concepts when this wouldn't have resulted in using an obscure word. We also made some choices to avoid having duplicate/competing concepts, i.e.\ \textit{boat} was translated as \textit{barca} and \textit{ship} as \textit{barco}. Further, we tried to match the intended use in English, i.e. translated \textit{sledgehammer} to \textit{almádena} rather than more generic term in Spanish \textit{mazo} as heavy metal version is more standard in English. Otherwise we tried to use more generic options. A variety of resources were used to aid this including bilingual dictionaries, Wikipedia, and RAE (Real Academia Española). Despite our best efforts to maintain as many concept-trait pairs as possible, certain concepts just don't work in Spanish, typically many to one translations, e.g. \textit{dove} translates to \textit{paloma} which also means normal mangy pigeons. A more common issue was the tendency to use multi-word expressions in Spanish for certain concepts, such as \textit{goldfish} (\textit{pez dorado}) and \textit{escalator} (\textit{escalera mecánica}) with no single-word alternatives. The statistics resulting to the trait subsets for {\mcrae}-ES are shown in the bottom section of Table \ref{tab:dateset}.

\parados{\norms-EN} To make our experiments more robust, we also used the {\norms} dataset. In order to use this dataset, we manually classified features in this dataset based on the subset from our {\mcrae} trait dataset. First, we cut the features in {\norms} that occurred less than 10 times and then took the set of remaining features and classified them as one of the five subsets and then automatically cast each concept-trait pair into their respective subset. We manually checked to see if any features not used had been erroneously omitted due to annotation issues and folded those features into the relative subsets. This entailed adding \texttt{is-liquid} and \texttt{is-furry} to the tactile subset after some consideration (with \texttt{is-furry} subsequently being removed due to the minimum frequency cut after removing duplicates). The resulting subsets had duplicate concepts removed and then a minimum frequency cut on the remaining features of 10. The statistics of the resulting subsets can be seen in the middle section of Table \ref{tab:dateset} with the number of new unique concepts added to each subset shown in parenthesis in the concept count (N$_C$) column.

\subsection{Corpora}\label{seccorpora}
\begin{table}[t!]
    \centering
    \small
    \renewcommand{\arraystretch}{1.1}
    \begin{tabular}{p{1.6em}lcc}
\toprule
& Corpus  & Sentences & Tokens \\
\midrule
\parbox[t]{5mm}{\multirow{3}{*}{\rotatebox[origin=c]{90}{\textbf{English}}}} &UMBC  & 135M & 3.4B  \\
&Wiki  & 114M & 2.5B \\
&Wee-Wiki & 71M & 1.6B \\
\midrule
\parbox[t]{5mm}{\multirow{3}{*}{\rotatebox[origin=c]{90}{\textbf{Spanish}}}}&ES1B  & 62M & 1.4B \\
&Wiki  & 28M & 0.6B \\
&Wee-Wiki & 19M & 0.4B \\
\bottomrule
    \end{tabular}
    \caption{Basic statistics of corpora used.}
    \label{tab:corpus-stats}
\end{table}
For the statistics of the corpora used see Table \ref{tab:corpus-stats}.

\parados{UMBC} The University of Maryland, Baltimore County (UMBC) webbase corpus is the resulting collection of paragraphs from a webcrawl in 2007 over millions of webpages \cite{han-etal-2013-umbc}.

\parados{ES1B} The Spanish Billion Words Corpus (ES1B) is a collection of unannotated sentences takens from the web which span difference sources from Europarl to books. It also include data from a Wikipedia dump from 2015, so has some crossover with the Spanish Wikipedia corpus \cite{cardellinoSBWCE}.

\parados{Wiki}
We used English Wikipedia dump from 1st October 2021 and Spanish Wikipedia dump from 1st January 2022. They were extracted and cleaned using the WikiExtractor tool from \citet{Wikiextractor2015}. This left document ID HTML tags in the data which we removed with a simple heuristic. 

\parados{Wee-Wiki}
Similar to the standard pre-processing of the Wikipedia data, but we also cut articles with very little views as these tend to be stub articles and automatically generated articles. The idea behind this is to cultivate a \textit{cleaner} and more natural version of the data. We used Wikipedia's official viewing statistics for 1st December 2021.\footnote{\url{https://dumps.wikimedia.org/other/pageviews/2021/2021-12/}} Articles with less than 10 views were removed.

\begin{table*}
    \centering
    \small
    \tabcolsep=.1cm
    \begin{tabular}{cl rrr} 
    \toprule
    & & \multicolumn{3}{c}{UMBC} \\
    & & \multicolumn{3}{c}{instances removed}\\
    & trait type & \multicolumn{1}{c}{sentence} & \multicolumn{1}{c}{window} & \multicolumn{1}{c}{syntactic} \\
    \midrule
     \parbox[t]{2.5mm}{\multirow{5}{*}{\rotatebox[origin=c]{90}{\textbf{{\mcrae}}}}} & colour & 76,800 & 70,159 & 8,974 \\
 & components & 33,284 & 23,347 & 9,745 \\
 & material & 28,061 & 19,171 & 6,030 \\
 & size \& shape & 104,478 & 68,697 & 18,213\\
 & tactile & 18,881 & 13,845 & 4,632\\
 \midrule
 \parbox[t]{2.5mm}{\multirow{5}{*}{\rotatebox[origin=c]{90}{\textbf{{\norms}}}}} & colour & 25,106 & 18,737 & 7,422\\
 & components & 5,270 & 3,793 & 1,291\\
 & material & 51,898 & 34,484 & 12,150\\
 & size \& shape & 105,895 & 68,162 & 18,372\\
 & tactile & 17,965 & 13,040 & 4,264\\
 \bottomrule
    \end{tabular}
    \hfill
    \begin{tabular}{rrr} 
    \toprule
    \multicolumn{3}{c}{Wiki} \\
    \multicolumn{3}{c}{instances removed}\\
    \multicolumn{1}{c}{sentence} & \multicolumn{1}{c}{window} & \multicolumn{1}{c}{syntactic}\\
    \midrule
    105,614 & 97,728 & 13,397\\
    22,307 & 15,500 & 5,915\\
    29,695 & 20,477 & 5,771\\
    131,165 & 88,453 & 26,612\\
    14,437 & 10,658 & 3,657\\
    \midrule
    26,378 & 19,824 & 8,360 \\
    4,463 & 3,110 & 1,005\\
    30,916 & 20441 & 7338\\
    117,210 & 79,041 & 22,812\\
    13,683 & 10,156 & 3,533\\
    \bottomrule
    \end{tabular}
    \hfill
    \begin{tabular}{rrr} 
        \toprule
    \multicolumn{3}{c}{Wee-Wiki} \\
    \multicolumn{3}{c}{instances removed}\\
    \multicolumn{1}{c}{sentence} & \multicolumn{1}{c}{window} & \multicolumn{1}{c}{syntactic}\\
    \midrule
    70,194 & 64,594 & 9,083\\
    15,553 & 10,987 & 4,544\\
    21,239 & 14,669 & 4,431\\
    90,280 & 60,516 & 17,452\\
    11,413 & 8,529 & 2,981\\
    \midrule
    19,561 & 14,777 & 6,581\\
    3,637 & 2,483 & 766\\
    21,051 & 13,823 & 4,694\\
    83,329 & 55,933 & 15,814\\
    11,048 & 8,307 & 2,929\\
    \bottomrule
    \end{tabular}
    \caption{Total instances removed and replaced for English Corpora (UMBC, Wiki, Wee-Wiki) for each dataset ({\mcrae} and {\norms}) by trait type and removal method (sentence, window, and syntactic as described in \S\ref{secremoving}).}
    \label{tab:removal_stats_en}
\end{table*}
\begin{table*}
    \centering
    \small
    \tabcolsep=.1cm
    \begin{tabular}{cl rrr} 
    \toprule
    & & \multicolumn{3}{c}{ES1B} \\
    & & \multicolumn{3}{c}{instances removed}\\
    & trait type &\multicolumn{1}{c}{sentence} & \multicolumn{1}{c}{window} & \multicolumn{1}{c}{syntactic}\\
    \midrule
     \parbox[t]{2.5mm}{\multirow{5}{*}{\rotatebox[origin=c]{90}{\textbf{{\mcrae}}}}} & colour &  31,267 & 25,121 & 208\\
 & components & 11,855 & 7,680 & 1,551\\
 & material & 8,473 & 6,087 & 1,344 \\
 & size \& shape & 34,416 & 19,276 & 248\\
 & tactile & 3,508 & 2,404 & 185\\
 \bottomrule
    \end{tabular}
    \hfill
    \begin{tabular}{rrr} 
    \toprule
    \multicolumn{3}{c}{Wiki} \\
    \multicolumn{3}{c}{instances removed}\\
   \multicolumn{1}{c}{sentence} & \multicolumn{1}{c}{window} & \multicolumn{1}{c}{syntactic}\\
    \midrule
    19,424 & 15,804 & 2,729\\
    6,628 & 4,048 & 1,873\\
    6,704 & 4,698 & 2,200\\
    23,224 & 13,513 & 4,001\\
    2,459 & 1,743 & 782\\
    \bottomrule
    \end{tabular}
    \hfill
    \begin{tabular}{rrr} 
        \toprule
    \multicolumn{3}{c}{Wee-Wiki} \\
    \multicolumn{3}{c}{instances removed}\\
   \multicolumn{1}{c}{sentence} & \multicolumn{1}{c}{window} & \multicolumn{1}{c}{syntactic}\\
    \midrule
    12,473 & 10,129 & 1,836\\
    4,318 & 2,716 & 1,317\\
    4,353 & 3,091 & 1,501\\
    15,584 & 9,157 & 2,798\\
    1,787 & 1,291 & 584\\
    \bottomrule
    \end{tabular}
    \caption{Total instances removed and replaced for each Spanish Corpora (ES1B, Wiki, Wee-Wiki) for the {\mcrae} dataset broken down by trait type and removal method (sentence, window, and syntactic as described in \S\ref{secremoving}).}
    \label{tab:removal_stats_es}
\end{table*}

\begin{figure}[t!]
    \centering
\begin{dependency}[edge style={black!80, thick},label style={fill=black!5},edge slant=7]
\begin{deptext}[column sep=1.20em,ampersand replacement=\^,font=\footnotesize]
  Cerró \^ la \^ \textbf{\textcolor{depblue}{puerta}} \^ de \^ el \^ \textbf{\textcolor{depred}{granero}} \\
  \end{deptext}
\depedge{3}{2}{\textsc{det}}
\depedge{1}{3}{\textsc{obj}}
\depedge{6}{4}{\textsc{case}}
\depedge{6}{5}{\textsc{det}}
\depedge[edge style={depblue!90, thick},label style={fill=depblue!25}]{3}{6}{\textsc{nmod}}
\deproot[edge unit distance=2.5ex]{1}{\textsc{root}}
\end{dependency}
\raggedright\small{Original text: \textit{Cerró la puerta del granero}}\\
\raggedright\small{English: \textit{She/he closed the barn door}}
    \caption{\textit{granero} (highlighted in red) is a concept in {\mcrae}-ES with a component trait of \textit{puerta} (highlighted in blue). In the example here they are linked by an \textit{nmod} edge (highlighted in blue). For the syntactic removal method this sentence would be removed.}
    \label{fig:syntax}
\end{figure}
\subsection{Removing co-occurrences}
\label{secremoving}

We used 3 methods to remove co-occurrences with different levels of granularity to find co-occurrences. The first step in the process was to segment the corpora by sentence and to lemmatise the tokens.  This was done using the spaCy library and the corresponding pre-trained models for English and Spanish \cite{spacy}. We used lemmas to handle gender of adjectives and nouns in Spanish and for plural forms in both languages. The segmented version of each corpus was then split into two separate corpora with 80\% of the sentences in the first, which were used as the standard corpora in our experiments, and with 20\%, which were used as reserves for replacing sentence with co-occurrences when creating input data without co-occurrences. When an instance was removed based on the criteria specified below, a random sentence was selected from the reserves, so as to balance the total number of sentences in each set.\footnote{\citet{chiang-etal-2020-understanding} observed only a small difference when using this methodology and when using one where instances were replaced with sentences containing the relative concepts (and as is shown in \S\ref{sec:results} this holds for our work).} The resulting number of instances removed is shown in Table \ref{tab:removal_stats_en} (English) and in Table \ref{tab:removal_stats_es} (Spanish).

\parados{Sentence}
The simplest method used was to merely remove any sentence where a concept and its corresponding trait was observed. The lemmatised version of the data was used to search for co-occurrences to be more thorough, especially with respect to the Spanish data. This entails using the lemmatised version of the concepts and traits to match them in the lemmatised instances in the data. This was done independently for each trait type. 

\parados{Window}
The second method used removed instances when the concept and its relative trait occurred within a given window, again using lemmatised forms. The window size used was 10 to match the size used during the training of the embeddings.

\parados{Syntactic}
Finally, used the Stanza library and the corresponding pre-trained models available for English and Spanish to parse the instances where a concept and its relative trait occurred \cite{qi2020stanza}. If an edge between the concept and the trait was predicted after finding a co-occurrence using the lemmas, this was removed, otherwise the instance was left. This method tests whether co-occurrences which are syntactically related are more impactful than haphazard co-occurrences. An example is shown in Figure \ref{fig:syntax}.

\subsection{Word embeddings}\label{secembeddings}
The models used to evaluate the impact of co-occurrences were trained using the Gensim library \cite{rehurek_lrec}. We used CBOW Word2Vec embedding models \cite{Mikolov2013EfficientEO} as they are quicker to train than skip-gram models which was paramount considering the number of models that were required. Further, \citet{chiang-etal-2020-understanding} found no significant differences between CBOW and Skip-gram models with respect to the differences observed in analogy completion between models trained with and without co-occurrences. We used the default hyperparameters in Gensim except for embedding size which was set to 300 and window size which was set to 10, i.e. the same settings from \citet{chiang-etal-2020-understanding}. For each trait-type and for each corpus a model was trained on the data containing co-occurrences (\textbf{with} or \textbf{w/} in tables) and the data not containing co-occurrences (\textbf{without} or \textbf{w/o} in tables). We trained multiple models for the data including co-occurrences --- once per trait type --- giving us a robust measurement of those models' performance. This means that results for each \textbf{with} for each trait type across the extraction methods are trained on the same data and are reported to show the variation seen training models on the same data.\footnote{Variation could also be due to slightly different datasets if \textbf{without} data doesn't contain any occurrences of a concept.}

\subsection{Classifiers}\label{secclassifiers}
Trait-based relational knowledge was evaluated by casting it as a classification problem. 

\begin{table*}
    \centering
    \small
    \tabcolsep=.055cm
    \begin{tabular}{llccp{0.5em}ccp{0.5em}cc}
    \toprule
    & & \multicolumn{8}{c}{UMBC} \\
    \cmidrule{3-10} 
    &  & \multicolumn{2}{c}{sentence} && \multicolumn{2}{c}{window} && \multicolumn{2}{c}{syntactic} \\
    & trait type & w/ & w/o && w/ & w/o && w/ & w/o \\
    \midrule
        \parbox[t]{2.5mm}{\multirow{5}{*}{\rotatebox[origin=c]{90}{\textbf{{\mcrae}}}}} & colour & 0.35 & 0.35 & & 0.34 & 0.34 & & \textbf{0.36} & 0.35 \\
             & components & \textbf{0.82} & 0.81 & & \textbf{0.81} & 0.80 & & \textbf{0.82} & 0.81 \\
             & materials & 0.65 & \textbf{0.69} & & \textbf{0.67} & 0.65 & & 0.67 & \textbf{0.68} \\
             & size \& shape & \textbf{0.57} & 0.53 & & 0.55 & \textbf{0.58} & & 0.54 & \textbf{0.58} \\
             & tactile & 0.61 & \textbf{0.62} & & \textbf{0.64} & 0.60 & & \textbf{0.65} & 0.64 \\
             \midrule
             \parbox[t]{2.5mm}{\multirow{5}{*}{\rotatebox[origin=c]{90}{\textbf{{\norms}}}}} & colour & \textbf{0.40} & 0.38 & & 0.41 & 0.41 & & 0.38 & \textbf{0.40} \\
             & components & 0.89 & 0.89 & & 0.89 & 0.89 & & 0.89 & 0.89 \\
             & materials & \textbf{0.88} & 0.87 & & \textbf{0.87} & 0.85 & & 0.87 & \textbf{0.88} \\
             & size \& shape & \textbf{0.59} & 0.57 & & 0.58 & \textbf{0.60} & & \textbf{0.61} & 0.58 \\
             & tactile & 0.69 & \textbf{0.72} & & \textbf{0.68} & 0.66 & & \textbf{0.70} & 0.66 \\
             \bottomrule
    \end{tabular}
    \hfill
        \begin{tabular}{ccp{0.5em}ccp{0.5em}cc}
    \toprule
     \multicolumn{8}{c}{Wiki} \\
     \cmidrule{1-8} 
    \multicolumn{2}{c}{sentence} && \multicolumn{2}{c}{window} && \multicolumn{2}{c}{syntactic} \\
     w/ & w/o && w/ & w/o && w/ & w/o \\
    \midrule
         \textbf{0.38} & 0.36 & & \textbf{0.38} & 0.30 & & \textbf{0.41} & 0.36 \\
0.78 & \textbf{0.80} & & 0.75 & \textbf{0.77} & & \textbf{0.79} & 0.76 \\
\textbf{0.68} & 0.67 & & 0.65 & \textbf{0.69} & & 0.65 & \textbf{0.67} \\
\textbf{0.60} & 0.58 & & \textbf{0.58} & 0.56 & & 0.56 & \textbf{0.61} \\
0.54 & \textbf{0.55} & & 0.56 & \textbf{0.59} & & \textbf{0.58} & 0.55 \\
             \midrule
        0.39 & 0.39 & & \textbf{0.41} & 0.39 & & \textbf{0.44} & 0.40 \\
0.91 & 0.91 & & 0.89 & \textbf{0.91} & & 0.91 & 0.91 \\
\textbf{0.86} & 0.84 & & 0.85 & 0.85 & & 0.86 & 0.86 \\
0.59 & 0.59 & & \textbf{0.62} & 0.57 & & 0.59 & \textbf{0.61} \\
0.61 & \textbf{0.65} & & 0.63 & 0.63 & & 0.65 & \textbf{0.67} \\
             \bottomrule
    \end{tabular}
    \hfill
     \begin{tabular}{ccp{0.5em}ccp{0.5em}cc}
    \toprule
     \multicolumn{8}{c}{Wee-Wiki} \\
     \cmidrule{1-8} 
    \multicolumn{2}{c}{sentence} && \multicolumn{2}{c}{window} && \multicolumn{2}{c}{syntactic} \\
     w/ & w/o && w/ & w/o && w/ & w/o \\
    \midrule
         \textbf{0.39} & 0.38 & & \textbf{0.39} & 0.38 & & \textbf{0.41} & 0.35 \\
\textbf{0.75} & 0.74 & & 0.75 & \textbf{0.79} & & 0.77 & 0.77 \\
\textbf{0.71} & 0.65 & & \textbf{0.67} & 0.66 & & 0.67 & \textbf{0.68} \\
\textbf{0.58} & 0.56 & & \textbf{0.59} & 0.56 & & \textbf{0.57} & 0.56 \\
0.50 & \textbf{0.51} & & 0.50 & \textbf{0.54} & & 0.51 & \textbf{0.55} \\
             \midrule
         \textbf{0.43} & 0.39 & & 0.37 & \textbf{0.39} & & 0.37 & \textbf{0.41} \\
0.89 & \textbf{0.91} & & 0.89 & \textbf{0.91} & & 0.94 & 0.94 \\
0.84 & 0.84 & & \textbf{0.83} & 0.82 & & \textbf{0.86} & 0.82 \\
\textbf{0.62} & 0.59 & & \textbf{0.58} & 0.55 & & \textbf{0.62} & 0.57 \\
0.60 & \textbf{0.61} & & 0.61 & 0.61 & & 0.63 & 0.63 \\
         \bottomrule
    \end{tabular}
    \caption{Multi-class SVM results for English corpora and datasets by trait type and extraction method for models trained on data with (\textbf{w/}) and without (\textbf{w/o}) co-occurrences. Average accuracy across 3-fold cross validation is reported with best performing model between paired \textbf{w/} and \textbf{w/o} models highlighted in bold.}
    \label{tab:en-multi}
\end{table*}

\begin{table*}
    \centering
    \small
    \tabcolsep=.055cm
    \begin{tabular}{llccp{0.5em}ccp{0.5em}cc}
    \toprule
    & & \multicolumn{8}{c}{ES1B} \\
    &  & \multicolumn{2}{c}{sentence} && \multicolumn{2}{c}{window} && \multicolumn{2}{c}{syntactic} \\
    & trait type & w/ & w/o && w/ & w/o && w/ & w/o \\
    \midrule
        \parbox[t]{2.5mm}{\multirow{5}{*}{\rotatebox[origin=c]{90}{\textbf{{\mcrae}}}}} & colour & 0.29 & \textbf{0.30} & & \textbf{0.33} & 0.31 & & \textbf{0.33} & 0.30 \\
             & components & \textbf{0.77} & 0.71 & & \textbf{0.81} & 0.77 & & \textbf{0.74} & 0.73 \\
             & materials & 0.63 & \textbf{0.67} & & \textbf{0.67} & 0.65 & & 0.66 & \textbf{0.67} \\
             & size \& shape & 0.50 & \textbf{0.52} & & 0.48 & \textbf{0.53} & & \textbf{0.46} & 0.45 \\
             & tactile & 0.54 & \textbf{0.58} & & \textbf{0.55} & 0.53 & & \textbf{0.55} & 0.53 \\
            \bottomrule
    \end{tabular}
    \hfill
        \begin{tabular}{ccp{0.5em}ccp{0.5em}cc}
    \toprule
     \multicolumn{8}{c}{Wiki} \\
    \multicolumn{2}{c}{sentence} && \multicolumn{2}{c}{window} && \multicolumn{2}{c}{syntactic} \\
     w/ & w/o && w/ & w/o && w/ & w/o \\
    \midrule
        \textbf{0.32} & 0.29 & & \textbf{0.34} & 0.32 & & \textbf{0.35} & 0.33 \\
0.71 & \textbf{0.75} & & 0.70 & \textbf{0.75} & & 0.72 & \textbf{0.74} \\
0.63 & \textbf{0.64} & & \textbf{0.70} & 0.63 & & 0.61 & \textbf{0.66} \\
\textbf{0.52} & 0.49 & & 0.49 & 0.49 & & 0.49 & \textbf{0.50} \\
\textbf{0.60} & 0.58 & & 0.60 & \textbf{0.62} & & \textbf{0.60} & 0.59 \\
            \bottomrule
    \end{tabular}
    \hfill
     \begin{tabular}{ccp{0.5em}ccp{0.5em}cc}
    \toprule
     \multicolumn{8}{c}{Wee-Wiki} \\
    \multicolumn{2}{c}{sentence} && \multicolumn{2}{c}{window} && \multicolumn{2}{c}{syntactic} \\
     w/ & w/o && w/ & w/o && w/ & w/o \\
    \midrule
         0.31 & \textbf{0.32} & & 0.30 & \textbf{0.31} & & \textbf{0.31} & 0.29 \\
\textbf{0.73} & 0.72 & & \textbf{0.71} & 0.66 & & 0.71 & 0.71 \\
0.59 & 0.59 & & 0.59 & \textbf{0.61} & & \textbf{0.63} & 0.59 \\
0.47 & \textbf{0.48} & & \textbf{0.49} & 0.48 & & 0.46 & \textbf{0.53} \\
\textbf{0.51} & 0.50 & & \textbf{0.52} & 0.51 & & \textbf{0.49} & 0.48 \\
            \bottomrule
    \end{tabular}
    \caption{Multi-class SVM results for Spanish corpora and datasets by trait type and extraction method for models trained on data with (\textbf{w/}) and without (\textbf{w/o}) co-occurrences. Average accuracy across 3-fold cross validation is reported with best performing model between paired \textbf{w/} and \textbf{w/o} models highlighted in bold.}
    \label{tab:es-multi}
\end{table*}

\parados{Multi-class}
First we used a multi-class evaluation. Using the datasets described in Section \ref{secdatasets}, given a concept (e.g. \textit{banana}), the task consisted of selecting the most appropriate trait for a given trait type (e.g. \textit{yellow} in the colour dataset). We used a support vector machine (SVM) as our classifier from the Scikit-learn library \cite{scikit-learn} with the word embeddings learned in the previous step as the only input. For each model we used 3-fold cross-validation and report the mean score across the splits.\footnote{The full results for each model can be found at \url{https://github.com/cardiffnlp/trait-relations-and-co-occurrences}, including the number of concepts and features used for each model's evaluation and the standard deviations which are all very small.} For each pair of models (i.e. with and without co-occurrences for a given trait-type and for a given corpus), we checked to see if concepts appeared in both semantic spaces. When a concept was missing in one or both, it was removed from the dataset for both, such that the comparison of results is robust between the two models we are interested in comparing, however, this was not common. It brought up an issue with \textit{orange} and \textit{naranja}, namely that it occurs as a concept and as trait, so that in our extraction method for sentence and window occurrences of these are always removed from the corpora and so were removed from the evaluation datasets. 

\parados{Binary}
We also use binary classification by exploiting the earlier findings suggesting that differences between embeddings can be used as a proxy to capture semantic relations \cite{Mikolov2013LinguisticRI,vylomova-etal-2016-take}. Again, we used SVM models, but this time the input features were the differences between concepts and their respective traits (i.e. $e_c - e_t$, where $e_c$ is the concept embedding and $e_t$ is the trait embedding) and the model predicted whether a pair was related or not. This required developing negative samples. This was done by randomly selecting words from the vocab space of the union of vocabs between each pair of model (i.e. with and without co-occurrences for a given trait type and a given corpus). These words then underwent a modicum of a control check by using lexical databases:
WordNet \cite{Fellbaum2000WordNetA} for English and the Multilingual Central Repository version 3.0 for Spanish \cite{Gonzalez-Agirre:Laparra:Rigau:2012} via the Natural Language Toolkit \cite{bird2009natural}. Once a word was randomly selected from the vocab space (excluding the concepts in the given dataset), the respective lexical database was checked to see if it contained the word and if so whether the synonyms associated with it were at least sometimes nouns (that is the synonym set of nouns contained at least one item). This was so that the selected word could in theory be something akin to a concept and not just gobbledygook. This procedure was done so the number of concepts in the negative sample set matched the number in the positive sample set (which had instances removed that didn't appear in one or both of the paired models similar to the multi-class setup). Then each randomly extracted negative \textit{concept} was ascribed a trait from the given trait space. Similar to the multi-class SVM setup, 3-fold cross-validation was used and the mean score across the splits is reported.\footnote{Full results for the binary classifier can be found at \url{https://github.com/cardiffnlp/trait-relations-and-co-occurrences}, including the number of instances for each model and the standard deviations.}

\begin{table*}
    \centering
    \small
    \tabcolsep=.055cm
    \begin{tabular}{llccp{0.5em}ccp{0.5em}cc}
    \toprule
    & & \multicolumn{8}{c}{UMBC} \\
    \cmidrule{3-10} 
    &  & \multicolumn{2}{c}{sentence} && \multicolumn{2}{c}{window} && \multicolumn{2}{c}{syntactic} \\
    & trait type & w/ & w/o && w/ & w/o && w/ & w/o \\
    \midrule
        \parbox[t]{2.5mm}{\multirow{5}{*}{\rotatebox[origin=c]{90}{\textbf{{\mcrae}}}}} & colour & \textbf{0.90} & 0.88 & & \textbf{0.88} & 0.86 & & 0.86 & 0.86 \\
             & components & \textbf{0.90} & 0.88 & & 0.90 & 0.90 & & 0.90 & 0.90 \\
             & materials & \textbf{0.93} & 0.92 & & 0.92 & 0.92 & & 0.90 & 0.90 \\
             & size \& shape & 0.88 & 0.88 & & 0.85 & 0.85 & & 0.85 & 0.85 \\
             & tactile & 0.89 & \textbf{0.90} & & 0.88 & 0.88 & & 0.86 & \textbf{0.88} \\
             \midrule
             \parbox[t]{2.5mm}{\multirow{5}{*}{\rotatebox[origin=c]{90}{\textbf{{\norms}}}}} & colour & 0.86 & 0.86 & & \textbf{0.84} & 0.83 & & 0.83 & 0.83 \\
             & components & 0.80 & \textbf{0.82} & & \textbf{0.87} & 0.77 & & 0.80 & 0.80 \\
             & materials & 0.84 & \textbf{0.85} & & \textbf{0.88} & 0.86 & & 0.88 & \textbf{0.91} \\
             & size \& shape & 0.84 & \textbf{0.87} & & 0.88 & \textbf{0.89} & & \textbf{0.87} & 0.86 \\
             & tactile & \textbf{0.87} & 0.84 & & \textbf{0.86} & 0.84 & & 0.84 & \textbf{0.86} \\
             \bottomrule
    \end{tabular}
    \hfill
        \begin{tabular}{ccp{0.5em}ccp{0.5em}cc}
    \toprule
     \multicolumn{8}{c}{Wiki} \\
     \cmidrule{1-8} 
    \multicolumn{2}{c}{sentence} && \multicolumn{2}{c}{window} && \multicolumn{2}{c}{syntactic} \\
     w/ & w/o && w/ & w/o && w/ & w/o \\
    \midrule
0.84 & \textbf{0.85} & & \textbf{0.88} & 0.86 & & 0.88 & 0.88 \\
\textbf{0.88} & 0.87 & & 0.87 & \textbf{0.88} & & 0.89 & \textbf{0.90} \\
\textbf{0.90} & 0.88 & & 0.88 & \textbf{0.89} & & 0.89 & 0.89 \\
0.86 & 0.86 & & 0.83 & 0.83 & & 0.84 & 0.84 \\
0.88 & 0.88 & & 0.82 & 0.82 & & 0.87 & \textbf{0.88} \\
             \midrule
        0.86 & 0.86 & & \textbf{0.85} & 0.83 & & \textbf{0.84} & 0.83 \\
0.90 & 0.90 & & \textbf{0.87} & 0.78 & & \textbf{0.86} & 0.84 \\
\textbf{0.89} & 0.87 & & 0.86 & \textbf{0.89} & & 0.85 & \textbf{0.87} \\
\textbf{0.88} & 0.86 & & 0.84 & 0.84 & & 0.85 & 0.85 \\
\textbf{0.83} & 0.82 & & 0.82 & \textbf{0.84} & & \textbf{0.86} & 0.84 \\
             \bottomrule
    \end{tabular}
    \hfill
     \begin{tabular}{ccp{0.5em}ccp{0.5em}cc}
    \toprule
     \multicolumn{8}{c}{Wee-Wiki} \\
     \cmidrule{1-8} 
    \multicolumn{2}{c}{sentence} && \multicolumn{2}{c}{window} && \multicolumn{2}{c}{syntactic} \\
     w/ & w/o && w/ & w/o && w/ & w/o \\
    \midrule
         0.89 & \textbf{0.90} & & 0.87 & 0.87 & & 0.87 & 0.87 \\
0.86 & 0.86 & & 0.92 & 0.92 & & 0.89 & 0.89 \\
\textbf{0.86} & 0.85 & & 0.88 & \textbf{0.89} & & \textbf{0.90} & 0.89 \\
\textbf{0.88} & 0.87 & & 0.87 & \textbf{0.88} & & 0.86 & 0.86 \\
\textbf{0.84} & 0.82 & & \textbf{0.84} & 0.81 & & \textbf{0.82} & 0.81 \\
             \midrule
       0.84 & 0.84 & & \textbf{0.87} & 0.86 & & \textbf{0.86} & 0.85 \\
0.86 & \textbf{0.87} & & \textbf{0.93} & 0.90 & & \textbf{0.84} & 0.83 \\
0.85 & \textbf{0.88} & & 0.85 & \textbf{0.88} & & 0.90 & \textbf{0.91} \\
0.88 & 0.88 & & \textbf{0.88} & 0.87 & & \textbf{0.87} & 0.85 \\
\textbf{0.78} & 0.77 & & 0.80 & \textbf{0.81} & & 0.86 & 0.86 \\
         \bottomrule
    \end{tabular}
    \caption{Binary SVM results for English corpora and datasets by trait type and extraction method for models trained on data with (\textbf{w/}) and without (\textbf{w/o}) co-occurrences. Average accuracy across 3-fold cross validation is reported with best performing model between paired \textbf{w/} and \textbf{w/o} models highlighted in bold.}
    \label{tab:en-binary}
\end{table*}

\begin{table*}
    \centering
    \small
    \tabcolsep=.055cm
    \begin{tabular}{llccp{0.5em}ccp{0.5em}cc}
    \toprule
    & & \multicolumn{8}{c}{ES1B} \\
    &  & \multicolumn{2}{c}{sentence} && \multicolumn{2}{c}{window} && \multicolumn{2}{c}{syntactic} \\
    & trait type & w/ & w/o && w/ & w/o && w/ & w/o \\
    \midrule
        \parbox[t]{2.5mm}{\multirow{5}{*}{\rotatebox[origin=c]{90}{\textbf{{\mcrae}}}}} & colour & 0.81 & \textbf{0.82} & & \textbf{0.85} & 0.83 & & \textbf{0.81} & 0.80 \\
             & components & \textbf{0.88} & 0.87 & & \textbf{0.83} & 0.80 & & \textbf{0.81} & 0.80 \\
             & materials & 0.81 & \textbf{0.82} & & 0.86 & 0.86 & & 0.84 & 0.84 \\
             & size \& shape & \textbf{0.82} & 0.81 & & \textbf{0.75} & 0.73 & & \textbf{0.76} & 0.74 \\
             & tactile & \textbf{0.72} & 0.71 & & 0.75 & \textbf{0.79} & & \textbf{0.81} & 0.78 \\
            \bottomrule
    \end{tabular}
    \hfill
        \begin{tabular}{ccp{0.5em}ccp{0.5em}cc}
    \toprule
     \multicolumn{8}{c}{Wiki} \\
    \multicolumn{2}{c}{sentence} && \multicolumn{2}{c}{window} && \multicolumn{2}{c}{syntactic} \\
     w/ & w/o && w/ & w/o && w/ & w/o \\
    \midrule
        \textbf{0.84} & 0.81 & & \textbf{0.81} & 0.78 & & \textbf{0.87} & 0.84 \\
0.86 & \textbf{0.89} & & 0.78 & 0.78 & & 0.79 & \textbf{0.82} \\
\textbf{0.78} & 0.76 & & 0.75 & 0.75 & & \textbf{0.70} & 0.67 \\
0.79 & \textbf{0.80} & & \textbf{0.76} & 0.75 & & \textbf{0.79} & 0.78 \\
\textbf{0.74} & 0.73 & & 0.71 & \textbf{0.72} & & 0.74 & \textbf{0.75} \\
         \bottomrule
    \end{tabular}
    \hfill
     \begin{tabular}{ccp{0.5em}ccp{0.5em}cc}
    \toprule
     \multicolumn{8}{c}{Wee-Wiki} \\
    \multicolumn{2}{c}{sentence} && \multicolumn{2}{c}{window} && \multicolumn{2}{c}{syntactic} \\
     w/ & w/o && w/ & w/o && w/ & w/o \\
    \midrule
        \textbf{0.83} & 0.81 & & \textbf{0.83} & 0.80 & & 0.79 & \textbf{0.82} \\
0.77 & \textbf{0.78} & & 0.82 & \textbf{0.84} & & \textbf{0.76} & 0.74 \\
0.75 & \textbf{0.80} & & \textbf{0.75} & 0.74 & & 0.74 & 0.74 \\
0.79 & 0.79 & & 0.75 & \textbf{0.78} & & 0.82 & \textbf{0.83} \\
\textbf{0.78} & 0.72 & & \textbf{0.77} & 0.75 & & 0.78 & \textbf{0.80} \\
         \bottomrule
    \end{tabular}
    \caption{Binary SVM results for Spanish corpora and datasets by trait type and extraction method for models trained on data with (\textbf{w/}) and without (\textbf{w/o}) co-occurrences. Average accuracy across 3-fold cross validation is reported with best performing model between paired \textbf{w/} and \textbf{w/o} models highlighted in bold.}
    \label{tab:es-binary}
\end{table*}

\section{Results}\label{sec:results}
\begin{figure}[t!]
    \centering
    \includegraphics[width=0.8\linewidth]{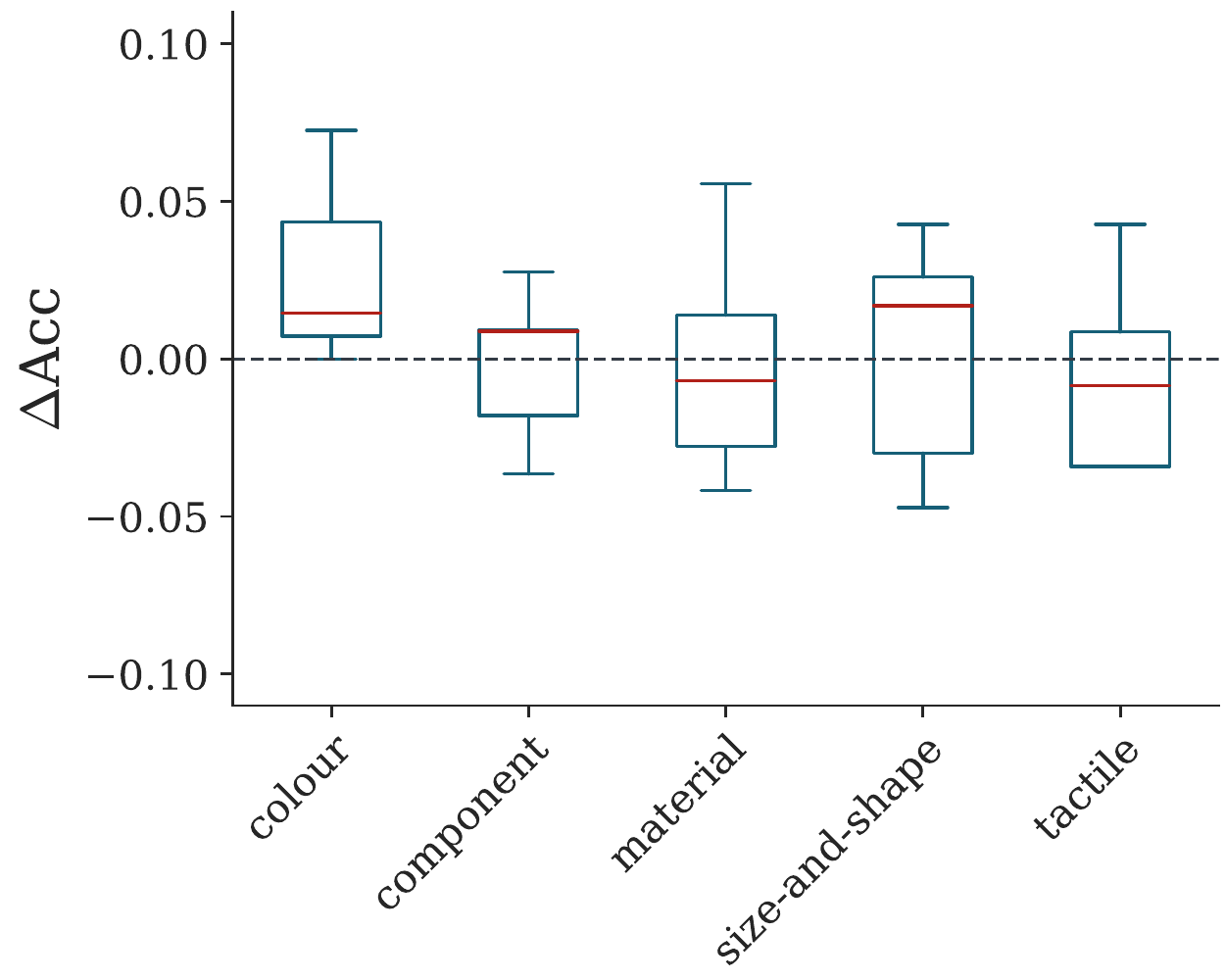}
    \caption{Distributions of delta accuracy ($\Delta$Acc) for corresponding pairs for each trait type in {\mcrae}-EN.}
    \label{fig:mcrae_en}
\end{figure}

\parados{Multi-class results} The results for the multi-class experiments can be seen in Table \ref{tab:en-multi} for the English corpora and in Table \ref{tab:es-multi} for the Spanish corpora. The highest performing model for each pair of models, i.e. with (\textbf{w/}) and without (\textbf{w/o}) co-occurrences is highlighted in bold for clarity. Across the board, it is clear that there is no consistent pattern as to whether a model trained with co-occurrences outperforms a model trained without them or vice versa. This holds for all three co-occurrence extraction techniques, for all trait types, for all datasets, and for all corpora across both languages. This is similar to the findings of \citet{chiang-etal-2020-understanding} where little effect was observed on analogy completion whether co-occurrences were included or not, however, a systemic decrease was observed in that context despite it being small. While there are some differences between some models, the differences that would be required to make claims of one model being superior to another are much larger than what are observed here as the experimental setup isn't robust enough to verify that a difference of 0.01-0.02 is significant or not. A visualisation of the differences between each corresponding with and without model for {\mcrae}-EN by trait type can be seen in Figure \ref{fig:mcrae_en} (equivalent visualisations for {\norms}-EN and {\mcrae}-ES are shown in Figure \ref{fig:norms_en} and \ref{fig:mcrae_es}, respectively, in Appendix \ref{sec:vis}). Figure \ref{fig:mcrae_en} does highlight a slight difference with respect to colour traits, where a modest increase in performance is seen on average when training the models with co-occurrences, however, this isn't consisted across corpora and datasets as this increase is not observed in Figures \ref{fig:norms_en} and \ref{fig:mcrae_es} in Appendix \ref{sec:vis}.

\parados{Binary results} The results from the binary classification experiments substantiate these findings. They can be seen in Table \ref{tab:en-binary} for English and in Table \ref{tab:es-binary} for Spanish. Again, no pattern emerges across the different experimental dimensions that would suggest the removal of co-occurrences has impacted a model's ability to predict whether a pair is related or not. The overall high performance on the binary classification experiment for both English and Spanish suggests these models manage to encode meaningful information about these trait relations. But how this emerges is not clear. The simplest explanation is that suitably accurate representations are learnt due to the amount of data, but it could be for any number of other reasons not investigated here.

\begin{figure}[b!]
    \centering
    \includegraphics[width=0.8\linewidth]{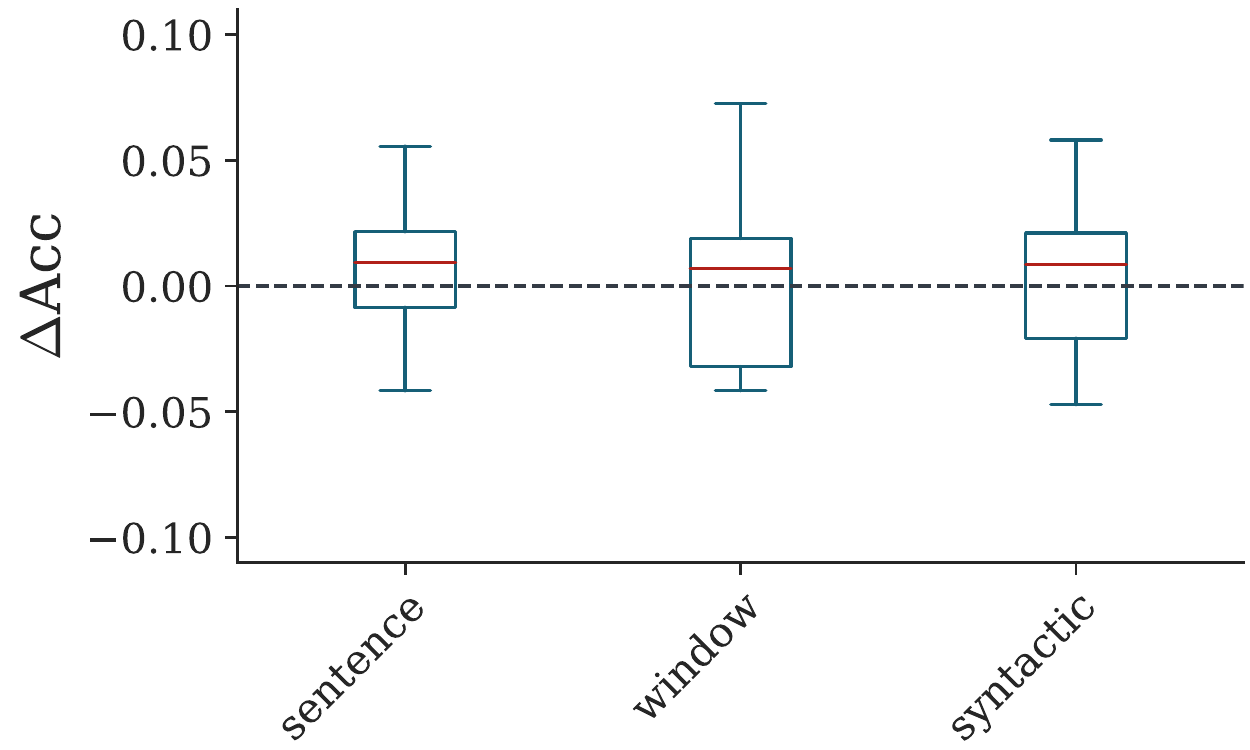}
    \caption{Distributions of delta accuracy ($\Delta$Acc) for pairs for each extraction method in {\mcrae}-EN.}
    \label{fig:mcrae_en_EM}
\end{figure}

\section{Discussion}
The results highlight some tentatively interesting patterns with respect to trait types. In both English and Spanish, models perform consistently well on component traits, although for {\norms} this turned out to be only over 2 traits, effectively casting it as a binary classification. Materials is the next consistently highest performing trait type across corpora and language with size \& shape and tactile not far behind for English, but with a bigger gap in Spanish. The performance on colour traits is low across all settings and languages. This doesn't appear to be based on the size of the trait subset, e.g. the component subset is one of the smaller sets, yet has high performance and the performance of the other trait types don't vary with respect to the number of instance and unique features. 

The number of removed sentences, as shown in Tables \ref{tab:removal_stats_en} and \ref{tab:removal_stats_es}, gives a vague indication of the occurrences of the concepts in the dataset and the occurrence of their traits with colour sentence removals  being the second highest for {\mcrae}-EN across all three English corpora, the third highest for {\norms}-EN, and the highest for {\mcrae}-ES  across all Spanish corpora. These rankings are consistent across extraction methods. Therefore, it is unlikely that the embeddings for the colours and the corresponding concepts (often concepts that occur in the other datasets) are somehow low quality due to low occurrences of these words. More likely is that the colour relation is more difficult than the other trait types as the other types are more tangible and more specific. Although this doesn't necessarily hold for size \& shape traits, specifically sizes which tend to be relative, e.g. in {\mcrae} a \textit{plane} can be \textit{large} (which it is, relative to most things) but so too can a \textit{bathtub}  (which it is, relative to a mouse or other such timorous beasties, but not relative to a house). However, size \& shape is consistently one of the traits that models perform worst on especially with {\norms}-EN and {\mcrae}-ES.

As a final note, the different extraction methods yield no differences when compared to one another. 
This can be observed clearly in Figure \ref{fig:mcrae_en_EM} in the main text and Figure \ref{fig:norms_en_EM} in Appendix \ref{sec:vis}. While the number of extracted instances using the syntactically related co-occurrences is very low and so difficult to draw any major conclusions, the number of sentence-based and window-based instances removed are quite high and are similar in magnitude. From this, we can deduce that the proximity of the words also doesn't have a major impact on the ability of a semantic space to encode relational knowledge. It could still be the case that if the data used to train models contained more syntactically related concept-trait pairs, they would encode \textit{more} relational knowledge, but it is clear that their absence doesn't result in the models losing what relational knowledge they can capture. Many questions remain on how these distributional models encode relational knowledge. We have merely presented results which \textit{do not} support the hypothesis that direct co-occurrence are the major signal for this process as related to trait-based relational knowledge.

\paragraph{Language models and wider impact of findings.}
Whether the results observed here for static embeddings would hold for PLMS isn't a given. While they are still based on the same distributional hypothesis and adopt statistical methods to encode salient features of language, they could potentially be more sensitive to the loss of co-occurrences in the training data. But this is an open research question that requires specific experimentation which has its own difficulties, i.e. prompting language models often includes lexical \textit{clues} which cloud our ability to say with any great certainty if they have captured some phenomenon or not, see \citet{kassner-schutze-2020-negated} for sensitivity of PLMs to mispriming). 

The results do suggest that merely increasing the amount of data used likely won't result in any major improvements in the ability of models to encode relational knowledge or commonsense knowledge more generally, which is attested to by recent work in \citet{Li2021DoLM}. Potentially, we need to look to more complex methods to augment NLP systems with commonsense knowledge potentially using multimodal systems, e.g. language models trained with visual cues as was done in \citet{paik-etal-2021-world} to offset reporting bias with respect to colours. Alternatively, we can focus on the linguistic input and consider how to add stronger signals in the data used to train NLP systems.

\section{Conclusion}
We have contributed to the emerging interest in how neural semantic models encode linguistic information, focusing on trait-based relational knowledge. We have extended findings which showed that co-occurrences of relational pairs didn't have a major impact on a model's ability to encode knowledge of analogies by complementing this analysis with an evaluation of trait-based relational knowledge. We extended the analysis to include different extraction methods to evaluate whether a more fine-grained approach would highlight any differences in performance and found that this is not the case. The work presented here also expands beyond English and includes results in Spanish which follow the same trend. Finally, we have cultivated a set of datasets for different trait types in both English and Spanish (based on {\mcrae} and {\norms}) which are available at \url{https://github.com/cardiffnlp/trait-concept-datasets}.

\section*{Acknowledgements}

Mark and Jose are supported by a UKRI Future Leaders Fellowship (MR/T042001/1).

\bibliography{sem}

\appendix
\section{{\mcrae}-EN trait subset extraction heuristics}\label{sec:heuristics}
Here we describe the full heuristics used to develop the trait-based subsets from {\mcrae} used in our experiments 

Some traits were trivial to extract. Colour relations were the simplest as they could be found using the feature classification in {\mcrae} of visual-colour. Component relations were shortlisted cutting on the WB feature classification (this is simply a classification of trait types where W and B refer to the practitioners who classified the concept-features pairs in unpublished work) in {\mcrae} using \texttt{external\_component} and \texttt{internal\_component} and then by extracting features beginning with \textit{has\_}. Similarly for material relations, the WB classification of \texttt{made\_of} was used. Some manual corrections were applied to the components to extend the number of instances in the dataset and to make certain traits fit our experimental setup better. This involved casting features such as \texttt{has-4-legs} and \texttt{has-4-wheels} as simply \texttt{has-legs} and \texttt{has-wheels}, respectively. The feature \texttt{made-of-material} was cut from the material subset, the feature \texttt{has-an-inside} from the components subset, and the features \texttt{is-colourful} and \texttt{different-colours} were removed from the colour subset. 

We then looked at the WB label \texttt{external\_surface\_property} (excluding features that fit into the colour, concept, or material subset) as this fit our desired trait-based feature space. The majority of concepts in this subset tended to have features relating to their shape or to their size, so we opted to use this pair (size \& shape) as another subset. This required manually removing features that didn't fit this trait-type, e.g.  \texttt{is-smelly}, \texttt{is-shiny}, and so on. In this process, a final possible subset of tactile-based traits became apparent which was cut using the BR feature classification (this is simply a classification of trait types from different practitioners than WB) \texttt{tactile} and then manually removing certain features which were more value judgements than traits, such as \texttt{is-comfortable} or \texttt{is-warm}.
\section{Visualisations of {\norms}-EN and {\mcrae}-ES results}\label{sec:vis}
\begin{figure}[htbp!]
    \centering
    \includegraphics[width=\linewidth]{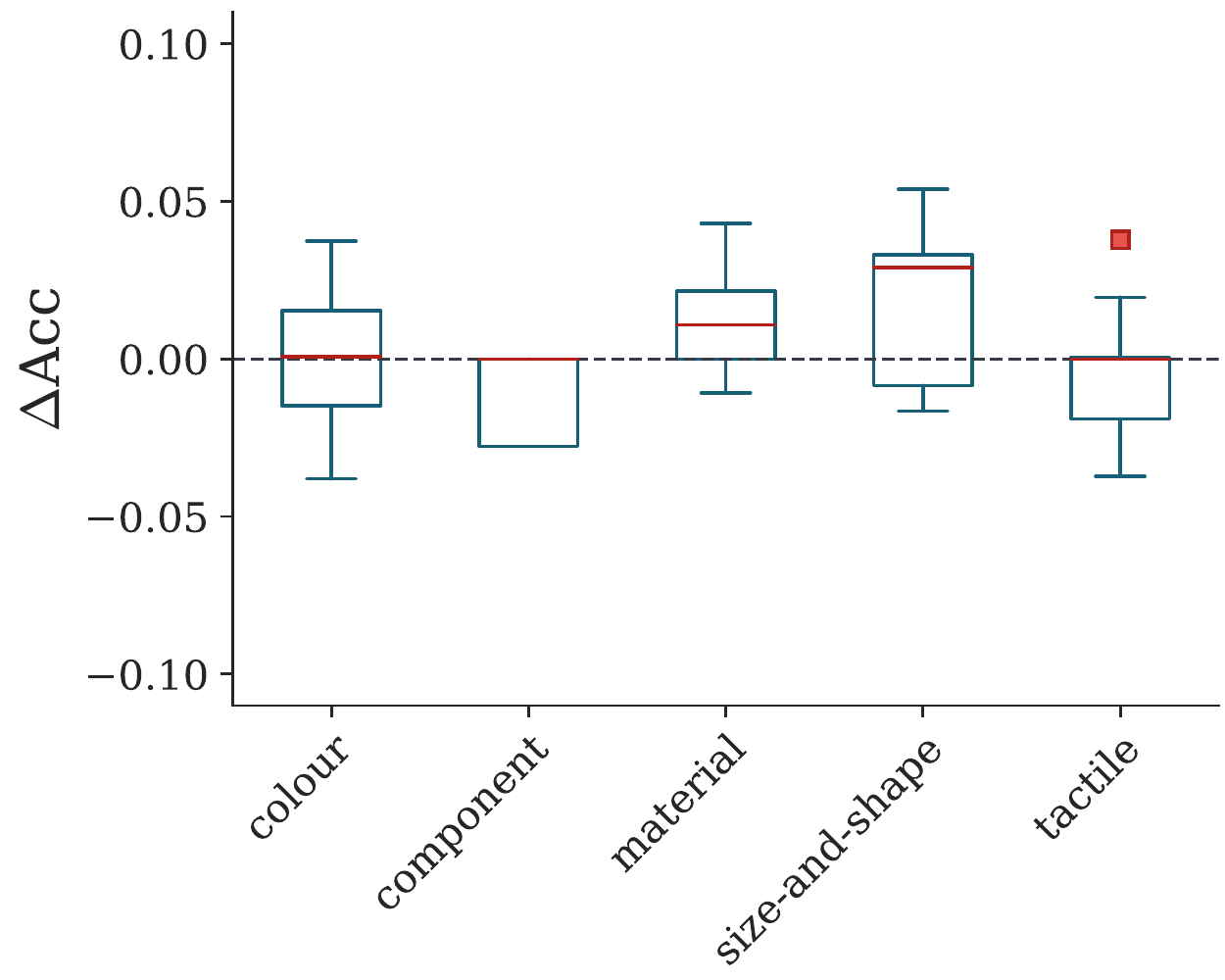}
    \caption{Distributions of delta accuracy ($\Delta$Acc) for corresponding pairs for each trait type in {\norms}-EN.}
    \label{fig:norms_en}
\end{figure}

\begin{figure}[htbp!]
    \centering
    \includegraphics[width=\linewidth]{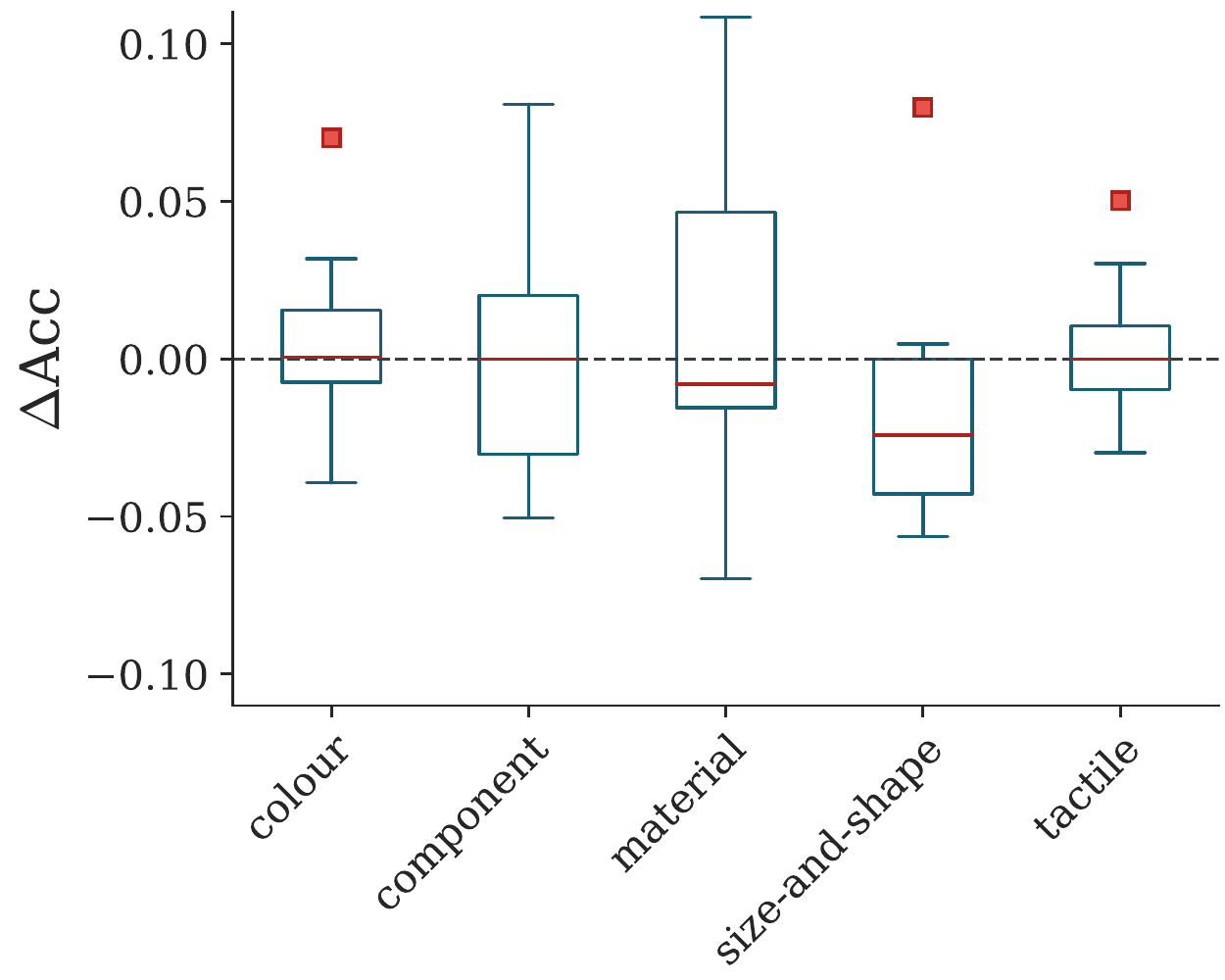}
    \caption{Distributions of delta accuracy ($\Delta$Acc) for corresponding pairs for each trait type in {\mcrae}-ES.}
    \label{fig:mcrae_es}
\end{figure}

\begin{figure}[H]
    \centering
    \includegraphics[width=\linewidth]{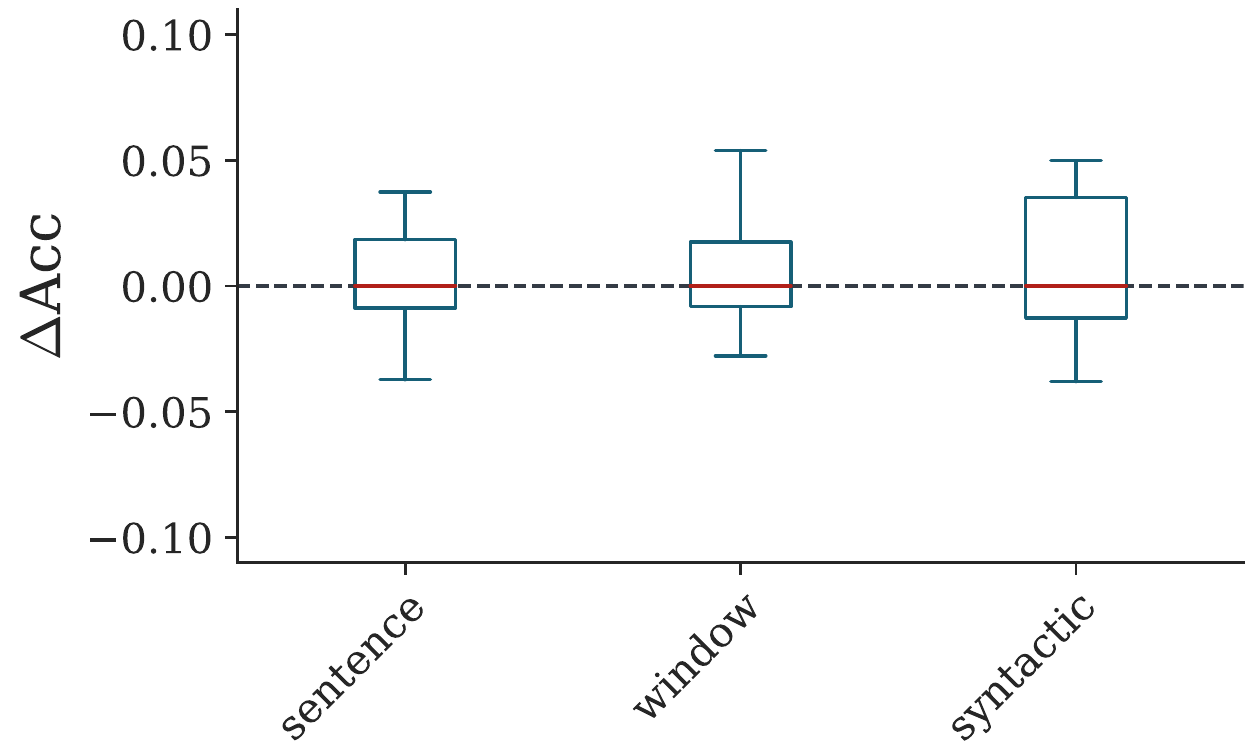}
    \caption{Distributions of delta accuracy ($\Delta$Acc) for corresponding pairs for extraction method in {\norms}-EN.}
    \label{fig:norms_en_EM}
\end{figure}

\begin{figure}[htbp!]
    \centering
    \includegraphics[width=\linewidth]{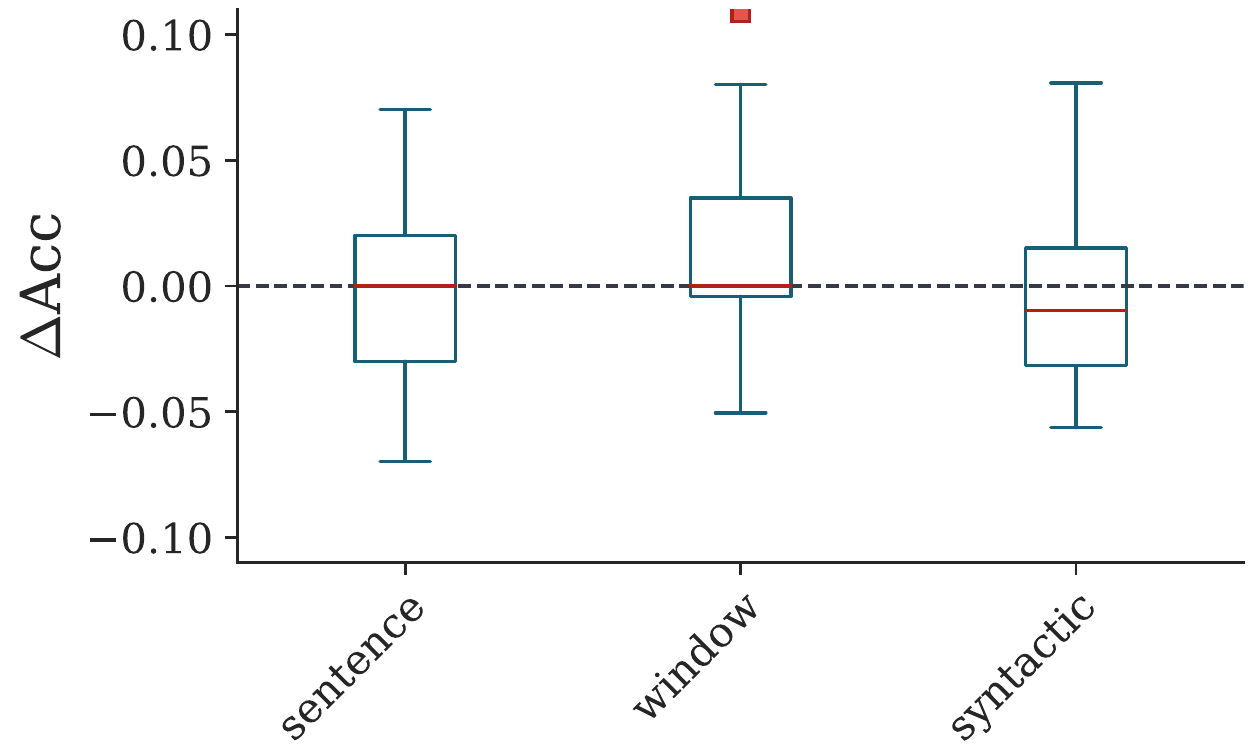}
    \caption{Distributions of delta accuracy ($\Delta$Acc) for corresponding pairs for each extraction method in {\mcrae}-ES.}
    \label{fig:mcrae_es_EM}
\end{figure}

\vspace{\fill}
\end{document}